\documentclass[11pt]{article}
\usepackage{times}
\usepackage{latexsym}
\usepackage[T1]{fontenc}
\usepackage[utf8]{inputenc}
\usepackage{microtype}
\usepackage{inconsolata}
\usepackage{graphicx}
\usepackage{amsmath,amssymb,booktabs}
\usepackage{xcolor}
\usepackage{array}
\usepackage[normalem]{ulem}
\usepackage[preprint]{acl}

\definecolor{revadd}{RGB}{0,84,180}

\newcommand{\dashsout}{\bgroup\markoverwith{\textcolor{gray}{\rule[0.5ex]{3pt}{0.5pt}}\hskip1.5pt}\ULon}

\title{When Should Decision Models Compare Candidates?}

\title{When Should Decision Models Compare Candidates?}

\author{
    \textbf{Fangzhi Li}\textsuperscript{1},
    \textbf{Quan Shi}\textsuperscript{2},
    \textbf{Wenting Yang}\textsuperscript{3},
    \textbf{Yueqiao Chen}\textsuperscript{4},
    \\
    \textbf{Jiayu Chang}\textsuperscript{5},
    \textbf{Shuting Zhao}\textsuperscript{3},
    \textbf{Zhaowei Xu}\textsuperscript{6},
    \\
    \textbf{Yizhao Chen}\textsuperscript{3},
    \textbf{Guanchu Wu}\textsuperscript{6},
    \textbf{Han Xu}\textsuperscript{7}
    \\[0.5em]
    \small \textsuperscript{1}William \& Mary
    \quad
    \textsuperscript{2}International School of Krakow
    \\
    \small\textsuperscript{3}University of California, San Diego
    \quad
    \textsuperscript{4}University of California, Davis
    \\
    \small
    \textsuperscript{5}Stanford University
    \quad
    \textsuperscript{6}University of California, Los Angeles
    \\
    \small
    \textsuperscript{7}University of Illinois Urbana-Champaign
    \\[0.5em]
    {\footnotesize
    \begin{tabular}{@{}ccc@{}}
    \texttt{fli05@wm.edu} &
    \texttt{qshi@iskrakow.org} &
    \texttt{wey023@ucsd.edu} \\[3pt]
    \texttt{yeqchen@ucdavis.edu} &
    \texttt{cjy1125@stanford.edu} &
    \texttt{shz106@ucsd.edu} \\[3pt]
    \texttt{zhaoweix@outlook.com} &
    \texttt{yic138@ucsd.edu} &
    \texttt{wuguanchu@g.ucla.edu} \\[3pt]
    \multicolumn{3}{c}{\texttt{hanxu8@illinois.edu}}
    \end{tabular}
    }
}

\begin{document}
\maketitle
\begin{abstract}
Language-model agents frequently select among
candidate actions, such as tools and web elements.
Should decision models score candidates independently
or allow their scores to depend on other candidates?
We investigate this question by comparing three
approaches: Pointwise scoring, Setwise scoring,
and Candidate Attention.
Across five agent-decision benchmarks,
Pointwise scoring remains competitive,
while Setwise scoring performs better on
some tool-selection and binary-decision tasks.
Candidate Attention does not consistently
improve upon Pointwise scoring, despite
allowing cross-candidate interaction.
Using two controlled synthetic tasks, we
further examine whether models learn to
use candidate-set information.
With targeted relational training,
Setwise models learn candidate dependencies
and generalize to shifted inputs,
while Candidate Attention shows weaker transfer.
Our results show that access to other
candidates does not guarantee effective
comparison, and that its benefits depend
on both task requirements and training.

\end{abstract}
\section{Introduction}
\label{sec:introduction}

Language-model agents frequently make decisions
among available alternatives, such as selecting
a tool, choosing a web element, or determining
which action best satisfies a request
~\citep{gu-etal-2023-dont,deng2023mind2web}.
These decisions can be handled by decision models
that assign scores to candidate actions and
select the highest-scoring one~\cite{typesafe2026jev}.
However, an important design choice remains:
should each candidate be evaluated
\textbf{independently}, or should its score
\textbf{depend on the other available candidates}?

\begin{figure}[t]
  \centering
  \includegraphics[width=\columnwidth]{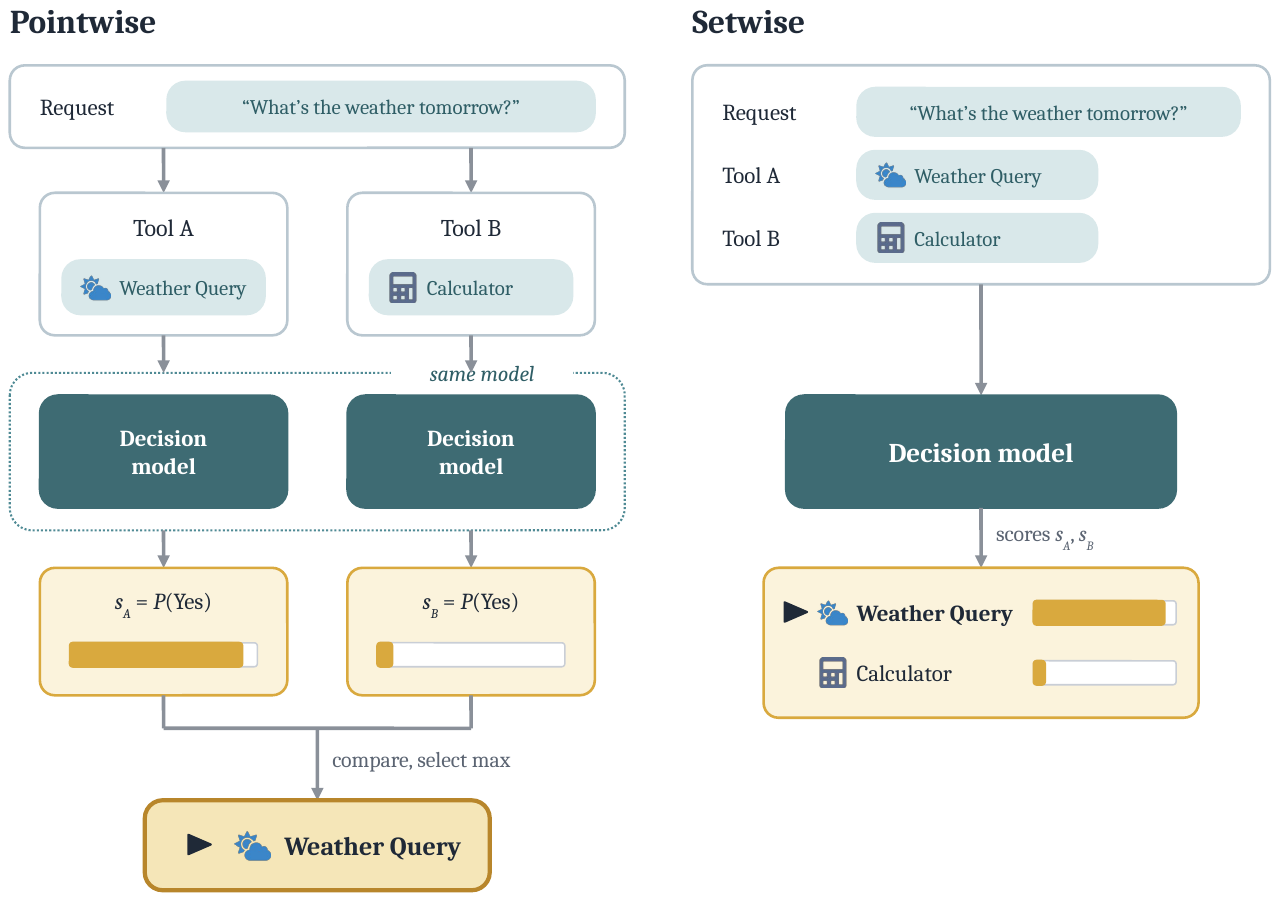}
  \caption{%
    Pointwise vs.\ Setwise candidate scoring.}
  \label{fig:formulations}
\end{figure}

\begin{figure*}[t]
  \centering
  \includegraphics[width=0.98\textwidth]{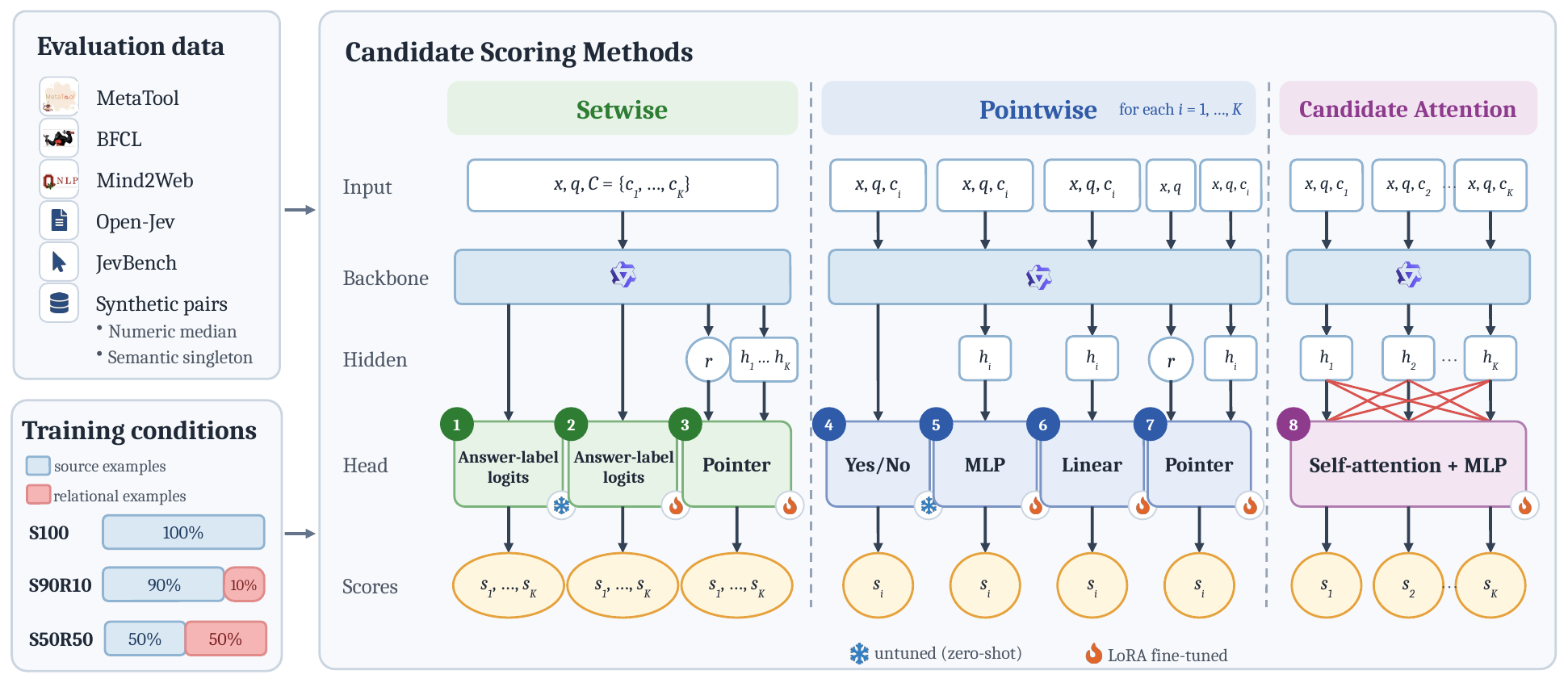}
\caption{\textbf{Method Overview.}
Left: Five evaluation benchmarks, two synthetic tasks,
and three training conditions.
Right: Eight scoring configurations grouped into
Setwise, Pointwise, and Candidate Attention.}
\label{fig:methods-overview}
\end{figure*}

These two approaches differ in what information
the decision model can use.
Illustrated in Figure~\ref{fig:formulations}, \textbf{Pointwise scoring} evaluates each candidate
using the shared context and that candidate alone.
\textbf{Setwise scoring} processes multiple candidates
together, allowing their representations or scores
to depend on one another.
Related distinctions between independent and
comparative scoring have been studied in
language-model ranking \citep{qin-etal-2024-large}.
Pointwise scoring supports independent candidate
encoding and reuse of intermediate representations,
while setwise scoring provides access to
relationships among alternatives.
Yet it remains unclear when this additional
information improves decision accuracy in practice.

We investigate three research questions (RQs):
\begin{enumerate}
    \item  When is Pointwise (independent) scoring sufficient?
    \item Can Candidate Attention match
Setwise scoring or improve upon Pointwise scoring?
    \item Which training conditions help, and do they generalize?
\end{enumerate}

To answer these questions, we compare three
scoring approaches: independent scoring (Pointwise),
joint candidate encoding (Setwise), and independent
encoding followed by cross-candidate attention (Candidate Attention), as shown in
Figure~\ref{fig:methods-overview}.
We evaluate these approaches on five
agent-decision benchmarks covering tool selection,
function calling, web element selection,
and other decision tasks.
We further construct two controlled relational
tasks to test whether models use information
about other candidates.
Finally, we vary the training mixture to examine
how relational supervision affects performance
and generalization.

Our experiments reveal three main findings.
(1) Pointwise scoring remains competitive,
outperforming Setwise scoring on some
candidate-selection and rating tasks.
(2) Candidate Attention does not consistently
improve upon Pointwise scoring.
(3) Access to other candidates does not
guarantee effective comparison:
with relational supervision, Setwise scoring
learns candidate dependencies, while
Candidate Attention remains less effective.

Together, these findings suggest that
the value of candidate interaction depends
on both task requirements and the model's
ability to use candidate information.

\section[Problem Formulation]{Problem Formulation} 
\label{sec:prob-formulation}
\subsection{Decision Setting}
\label{sec:decision-setting}

Let $x$ denote the state available to an agent, $q$ a decision question, and $C=\{c_1,\ldots,c_K\}$ the available candidates.
The model assigns each candidate $c_i$ a score
$s_i\in\mathbb{R}$, where higher scores indicate greater preference. The model then selects the candidate with the highest score: $c^* = \arg\max_{c_i \in C} s_i$. We distinguish two scoring formulations:
\vspace{1pt}
\begin{align*}
    \text{Independent:}\quad & s_i=f_\theta(x,q,c_i), \\
    \text{Interacting:}\quad & s_i=g_\theta(x,q,c_i,C).
\end{align*}

Independent scoring (Pointwise) evaluates each candidate $c_i$
using only $x$, $q$, and $c_i$, without access to other candidates
in $C$.
In contrast, interacting scoring (Setwise and Candidate Attention)
allows the score of $c_i$ to depend on other candidates in $C$,
either through joint encoding (Setwise) or through an attention
head over independently encoded candidate representations
(Candidate Attention).

Both formulations compare candidate scores during final selection. While softmax normalization couples the resulting probabilities across candidates in $C$,
it preserves the relative ranking of unchanged candidates under independent scoring. Our distinction therefore concerns whether computing $s_i$ can use information from other candidates, rather than how the scores are normalized or used for selection.

\begin{table*}[t]
\centering
\small
\setlength{\tabcolsep}{4pt}
\resizebox{\textwidth}{!}{%
\begin{tabular}{lllccl}
\toprule
Method
& \shortstack[l]{Candidate\\encoding}
& LM Backbone input
& \shortstack{LM Backbone\\fine-tuning}
& \shortstack{Additional\\head}
& Scoring rule \\
\midrule
1. Setwise LM (untuned)
& Joint & $(x,q,C)$
& No & No & Answer-label logits \\

2. Setwise LM (fine-tuned)
& Joint & $(x,q,C)$
& Yes & No & Answer-label logits \\

3. Setwise Pointer
& Joint & $(x,q,C)$
& Yes & Yes & Pointer head \\

4. Pointwise Yes/No
& Independent & Each $(x,q,c_i)$ separately
& No & No & Yes probability \\

5. Pointwise MLP
& Independent & Each $(x,q,c_i)$ separately
& Yes & Yes & Shared MLP \\

6. Pointwise Linear
& Independent & Each $(x,q,c_i)$ separately
& Yes & Yes & Shared linear head \\

7. Pointwise Pointer
& Independent & $(x,q)$ and each $(x,q,c_i)$ separately
& Yes & Yes & Pointer head \\

8. Candidate Attention
& Independent & Each $(x,q,c_i)$ separately
& Yes & Yes & Self-attention and MLP \\
\bottomrule
\end{tabular}%
}
\caption{Eight experimental
configurations. Candidate encoding describes how the
backbone processes candidates. An additional head is a trained module beyond the backbone's original language-model output layer.}
\label{tab:scoring_methods}
\end{table*}

\subsection{Candidate-Set Dependence}
The correct choice may depend on other candidates in $C$.
Consider two candidate sets $C$ and $C'$ with the same
state $x$, question $q$, and candidates $a$ and $b$.
Changing another candidate can reverse the correct choice:
\begin{equation*}
    y^*(x,q,C)=a,\qquad y^*(x,q,C')=b,
\end{equation*}
where $y^*$ denotes the correct choice.

For example, when selecting the median, the correct
choice changes from $5$ in $\{2,5,8\}$ to $2$ in
$\{2,5,1\}$, even though candidates $2$ and $5$
remain unchanged.

Under deterministic scoring and fixed tie-breaking,
independent scoring assigns the same scores to $a$
and $b$ in both sets. Their relative ranking cannot
change, so independent scoring cannot be correct
on both instances. In contrast, interacting scoring allows scores to change with $C$, making such reversals possible. Whether models learn this dependence is evaluated empirically.

We evaluate this limitation using paired synthetic tasks
in Section~\ref{sec:benchmarks}, where each example has
a counterpart whose correct answer is the other
unchanged candidate.

\section{Methods}
\label{sec:methods}

Building on the scoring formulations in
Section~\ref{sec:decision-setting},
we compare three approaches: Pointwise, Setwise,
and Candidate Attention, across eight configurations.
All methods use a language model (LM) as the
backbone of the decision model, converting input
contexts and candidate actions into scores.
They differ in how candidates are encoded, as summarized in
Table~\ref{tab:scoring_methods}.
Additional architecture and implementation
details are provided in Appendix~\ref{app:prompts}.

\paragraph{Setwise LM.}
The model receives the state $x$, question $q$, and
candidate set $C$ in a single prompt. Each candidate
$c_i$ is assigned a single-token answer label $t_i$
(e.g., A or B). We score each candidate using the
next-token logit of its label:
\begin{equation*}
    s_i = z_\theta(t_i \mid x,q,C),
\end{equation*}
where $z_\theta$ denotes the LM's next-token logit
at the answer position. We evaluate both the
untuned LM and a task-specific fine-tuned version.

\paragraph{Setwise Pointer.}
This method also encodes $(x,q,C)$ jointly.
We extract a query representation $r$ from the
final answer position and a representation $h_i$
from the last token of each candidate $c_i$.
A learned pointer head computes:
\begin{equation*}
    s_i = \frac{Q_\phi(r)^\top K_\phi(h_i)}{\sqrt{d}},
\end{equation*}
where $Q_\phi$ and $K_\phi$ share layer normalization
but use separate learned projections.
We train the pointer head jointly with the LM backbone.

\paragraph{Pointwise Yes/No.}
The model evaluates each candidate $c_i$ separately,
given only $(x,q,c_i)$, and predicts whether
$c_i$ is correct. Let $z_i^{\mathrm{Yes}}$ and $z_i^{\mathrm{No}}$
denote the next-token logits for Yes and No. The candidate score is the
normalized Yes probability:
\begin{equation*}
    s_i =
    \frac{\exp(z_i^{\mathrm{Yes}})}
    {\exp(z_i^{\mathrm{Yes}})+\exp(z_i^{\mathrm{No}})}.
\end{equation*}
This baseline uses the untuned LM without
task-specific training.

\paragraph{Pointwise MLP.}
We independently encode each candidate with
the shared state and question:
\begin{align*}
    h_i &= \operatorname{Enc}_\theta(x,q,c_i),\\
    s_i &= \operatorname{MLP}_\phi(h_i),
\end{align*}
where $h_i$ is the final hidden state and
$\operatorname{MLP}_\phi$ is a scoring head
shared across candidates. Neither encoding
nor scoring accesses other candidates.

\paragraph{Pointwise Linear.}
This method uses the same independent encoding
as Pointwise MLP but replaces the MLP with
a linear scoring head:
\begin{equation*}
    s_i = w^\top \operatorname{LN}(h_i)+b,
\end{equation*}
where $\operatorname{LN}$ denotes layer normalization,
and $w$ and $b$ are shared across candidates.
This variant tests whether a linear head is
sufficient for candidate scoring.

\paragraph{Pointwise Pointer.}
Unlike Setwise Pointer, this method obtains
the query representation without candidate
information and encodes each candidate separately:
\begin{align*}
    r &= \operatorname{Enc}_\theta(x,q),\\
    h_i &= \operatorname{Enc}_\theta(x,q,c_i),\\
    s_i &= \frac{Q_\phi(r)^\top K_\phi(h_i)}
    {\sqrt{d}}.
\end{align*}
Here, $r$ and $h_i$ are the final hidden states
of their respective prompts. We use the same
pointer-head architecture as Setwise Pointer,
with separately trained parameters.
Since neither representation accesses other
candidates, $s_i$ depends only on $(x,q,c_i)$.

\paragraph{Candidate Attention.}
We first encode each candidate independently,
as in Pointwise MLP, then allow their
representations to interact through an
attention-based scoring head:
\begin{equation*}
    (s_1,\ldots,s_K)
    = \operatorname{Attn}_\phi(h_1,\ldots,h_K).
\end{equation*}
The head projects the candidate representations,
applies 4-head self-attention, and scores
each updated representation with a shared MLP.
Unlike pointwise methods, each score can
depend on other candidates in $C$.

\begin{table*}[t]
\centering
\small
\setlength{\tabcolsep}{4pt}
\renewcommand{\arraystretch}{0.92}
\begin{tabular}{ll*{9}{r}}
\toprule
& & & \multicolumn{3}{c}{Setwise} & \multicolumn{4}{c}{Pointwise} & \\
\cmidrule(lr){4-6}\cmidrule(lr){7-10}
Dataset & Subtask & $n$ & LM (untuned) & LM (tuned) & Pointer & Yes/No & MLP & Linear & Pointer & Cand.\ Attn. \\
\midrule
MetaTool & Scenario & 1,450 & 12.62 & 84.48 & \textbf{84.97} & 45.66 & 82.28 & 80.48 & 80.41 & 80.48 \\
 & Similar & 802 & 0.00 & \textbf{76.68} & 75.94 & 51.50 & 74.19 & 74.81 & 74.19 & 74.31 \\
 & Reliability & 802 & \textbf{100.00} & 94.89 & 85.16 & 0.00 & 72.44 & 87.16 & 60.10 & 76.68 \\
 & Overall & 3,054 & 32.25 & \textbf{85.17} & 82.65 & 35.20 & 77.57 & 80.75 & 73.44 & 77.87 \\
\midrule
BFCL & Choice & 835 & 35.45 & \textbf{93.41} & 60.84 & 55.21 & 91.14 & 90.30 & 88.98 & 91.02 \\
 & Binary & 1,048 & \textbf{83.49} & 75.86 & 55.82 & 16.51 & 55.44 & 60.88 & 54.77 & 59.45 \\
 & Overall & 1,883 & 62.19 & \textbf{83.64} & 58.05 & 33.67 & 71.27 & 73.92 & 69.94 & 73.45 \\
\midrule
Mind2Web & Retrieved & 935 & 6.52 & \textbf{26.31} & 22.89 & 17.01 & 21.71 & 23.21 & 21.18 & 21.18 \\
$K{=}16$ & All & 1,312 & 4.65 & \textbf{18.75} & 16.31 & 12.12 & 15.47 & 16.54 & 15.09 & 15.09 \\
\midrule
Mind2Web & Retrieved & 1,127 & 2.57 & 8.52 & 15.26 & 9.67 & 15.08 & \textbf{16.42} & 12.95 & 15.17 \\
$K{=}64$ & All & 1,312 & 2.21 & 7.32 & 13.11 & 8.31 & 12.96 & \textbf{14.10} & 11.13 & 13.03 \\
\midrule
Open-Jev & Choice & 500 & 30.40 & 40.20 & 41.60 & 59.80 & \textbf{62.60} & 61.40 & 60.40 & 62.20 \\
 & Binary & 1,205 & 70.04 & 71.45 & \textbf{71.95} & 36.10 & 65.64 & 64.32 & 69.63 & 69.21 \\
 & Rating & 297 & 25.59 & 52.86 & 56.90 & 36.70 & 60.94 & \textbf{62.29} & 58.25 & 58.92 \\
 & Overall & 2,002 & 53.55 & 60.89 & 62.14 & 42.11 & 64.19 & 63.29 & 65.63 & \textbf{65.93} \\
\midrule
JevBench & Choice & 94 & 46.81 & 72.34 & 74.47 & 50.00 & \textbf{75.53} & 71.28 & 73.40 & 70.21 \\
 & Binary & 49 & 57.14 & \textbf{73.47} & 71.43 & 38.78 & 69.39 & 65.31 & 65.31 & 71.43 \\
 & Rating & 12 & 16.67 & 75.00 & 75.00 & 41.67 & \textbf{83.33} & \textbf{83.33} & \textbf{83.33} & \textbf{83.33} \\
 & Overall & 155 & 47.74 & 72.90 & 73.55 & 45.81 & \textbf{74.19} & 70.32 & 71.61 & 71.61 \\
\bottomrule
\end{tabular}
\caption[Accuracy on the five decision benchmarks]{Accuracy (\%) on the five decision benchmarks. All methods use Qwen3.5-2B-Base, fine-tuned with main training set S100. Setwise LM (untuned) and Pointwise Yes/No receive no task-specific fine-tuning.}
\label{tab:benchmark_results}
\end{table*}

\section{Experiments}
\label{sec:experiments}

To address the three research questions (RQs) introduced in
Section~\ref{sec:introduction}, we evaluate the scoring
methods on five existing agent-decision benchmarks and
two controlled synthetic tasks. The benchmarks test
generalization beyond the training data, while the
synthetic tasks test whether models can use
candidate-set information.

\subsection[Training Data and Conditions]{Training Data and Conditions}
\label{sec:experimental_setup}

We use two training pools: a source pool ($S$) and
a relational pool ($R$). The source pool contains
115,366 examples from the Strands Decider v19
data mixture\footnote{\url{https://github.com/strands-labs/strands-decider}},
covering classification, sentiment analysis, inference,
question answering, paraphrase identification,
rule-based reasoning, and generated decision tasks.
\footnote{
Sources include VitaminC, MNLI, WNLI, ContractNLI,
BoolQ, MuSiQue, PubMedQA, SST-5, Yelp, App Reviews,
CLINC150, BANKING77, AG News, DBpedia, Yahoo Topics,
PAWS, RuleTaker, and additional classification,
adequacy, and synthetic decision tasks.
}
The relational pool contains 6,000 synthetic examples,
equally divided between numeric median and semantic
singleton selection.

\paragraph{Training Conditions.}
We evaluate three conditions:
{\small
\begin{enumerate}
    \item \emph{S100}: Source examples only.
    \item \emph{S90R10}: 90\% source and 10\% relational examples.
    \item \emph{S50R50}: 50\% source and 50\% relational examples.
\end{enumerate}
}

The first three conditions use 3,606 optimizer updates
with a batch size of 32, while R100-short uses 500 updates.
Relational examples are resampled as needed.

\subsection{Experimental Design.}
We organize the experiments around three research
questions:

\noindent\textbf{RQ1: When is Pointwise scoring sufficient?}
We compare all methods on five external agent-decision
benchmarks under S100 and evaluate candidate-set
dependence using paired synthetic tasks.

\noindent\textbf{RQ2: Does Candidate Attention improve decision accuracy?}
We compare Setwise LM, Candidate Attention, and
Pointwise MLP using the same training examples
and number of updates. Controlled comparisons
further isolate the effect of candidate interaction.

\noindent\textbf{RQ3: Which training conditions help,
and do they generalize?}
We compare independent and interacting scoring across
S100, S90R10, S50R50.

\subsection{Evaluation Tasks and Benchmarks}
\label{sec:benchmarks}

\subsubsection{Existing Benchmarks}

We evaluate the scoring methods on five external
agent-decision benchmarks: BFCL, MetaTool, Open-Jev,
JevBench, and Mind2Web, using the same 9,156 evaluation
requests for each method.
These benchmarks test tool selection, function
selection, web element selection, and other agent
decisions. All evaluation examples are excluded
from the training pools.
We evaluate individual decisions from given contexts,
rather than complete sequences of agent actions.

\paragraph{Tool and Function Selection.}
We use MetaTool \citep{huang2024metatool} and an
adaptation of the Berkeley Function Calling
Leaderboard (BFCL) \citep{patil2025bfcl}.
MetaTool includes scenario-based tool selection
and selection among tools with similar descriptions.
These settings allow us to evaluate decisions
with different degrees of candidate overlap.
Our BFCL adaptation evaluates function selection
and whether the available functions can handle
a request.

\paragraph{Web element selection.}
We use Mind2Web \citep{deng2023mind2web} to evaluate
selection among candidate webpage elements.
The task instruction and previous actions form
the shared context, and each candidate is represented
by its element description.
We retain the official cross-task, cross-website,
and cross-domain test partitions.

We evaluate the same web interaction steps using
candidate pools of up to 16 or 64 retrieved elements.
We report two accuracy measures:
(1) retrieved accuracy, computed on steps where
the candidate pool contains at least one correct element;
and (2) overall accuracy, computed on all steps,
with retrieval failures counted as errors.
We do not add missing correct elements to the
retrieved candidate pools.

\paragraph{JEV Decision Tasks.}
Open-Jev and JevBench cover three decision types:
(1) selecting a candidate,
(2) predicting whether an action or function
is suitable for a request (Yes/No), and
(3) predicting a rating on an ordered scale.

\subsubsection{Controlled Synthetic Tasks}
\label{sec:synthetic_tasks}

We also construct two synthetic tasks to test
candidate-set dependence. Each example forms a pair
of decisions with the same state $x$, question $q$,
and two unchanged candidates. Replacing another
candidate reverses the correct choice between them. We also test candidate-order sensitivity using
canonical ordering and prediction averaging over
cyclic shifts. Details are provided in
Appendix~\ref{app:order}.

\paragraph{Numeric Median Selection.}
We construct synthetic tool descriptions with
distinct fees and ask the model to select the
tool with the median fee.
For example, given fees of $\{10,20,30\}$,
the correct choice is $20$.
Replacing $30$ with $5$ changes the correct
choice to $10$, while the other two fees remain
unchanged.

\paragraph{Semantic Singleton Selection.}
We ask the model to select the only tool whose
function does not repeat, while all other
functions appear twice with different wording.
For example, given ``Translate Japanese into
English,'' ``Merge several PDFs into one file,''
and ``Convert Japanese text to English,''
the PDF tool is correct.
Replacing the third tool with ``Combine PDF
files into a single PDF'' makes the first
translation tool correct.

\paragraph{Dataset Splits.}
Across both synthetic tasks, we use 6,000 training
examples (3,000 pairs), 600 validation examples
(300 pairs), and 1,200 test examples (600 pairs).
Both instances of each pair remain in the same
split, with no overlap between splits.
Candidate sets contain 3, 5, or 7 candidates.

\subsection{Model training.}

All fine-tuned methods use Qwen3.5-2B-Base
with LoRA (rank 16). We train the LoRA adapters
and any additional scoring heads while keeping
the original LM weights frozen.
We use the final checkpoint from each run.
Further implementation details are provided in
Appendix~\ref{app:impl}.

\subsection{Controlled Comparisons}
\label{sec:controlled}

Our main comparison is between Pointwise MLP and
Candidate Attention. Both use the same backbone
initialization, candidate prompts, training data,
and number of updates, with matched head capacity
(approximately 1.06M parameters).

Since backbones are fine-tuned separately,
we also compare both heads using identical frozen
candidate representations to better isolate the
effect of candidate interaction.
Further details are provided in
Appendix~\ref{app:controlled}.

\paragraph{Code and Data Availability.}
Our training and evaluation code, along with the
processed datasets and synthetic data generation
scripts, are available at
\url{https://anonymous.4open.science/r/decision-visibility-63FF}.

\section{Results}
\begin{figure*}[t]
    \centering
    \begin{minipage}[t]{0.32\textwidth}
        \centering
        \includegraphics[width=\linewidth]{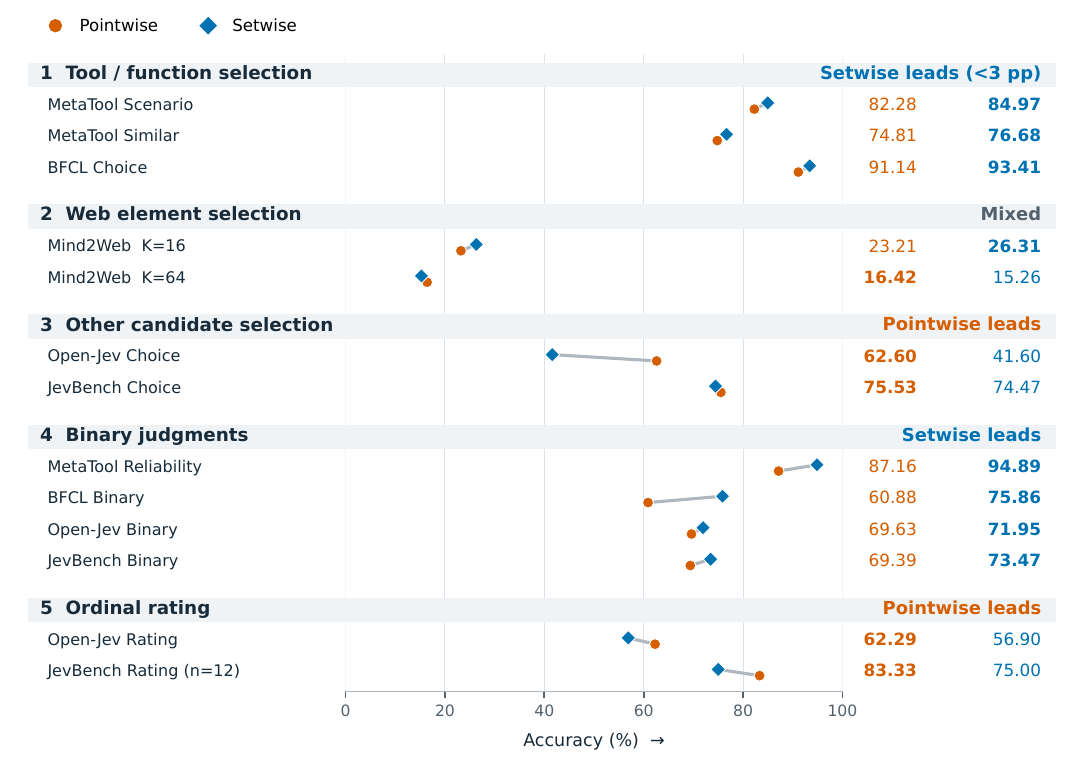}
        \par\smallskip
        {\small (a) RQ1: Pointwise vs.\ Setwise}
    \end{minipage}
    \hfill
    \begin{minipage}[t]{0.32\textwidth}
        \centering
        \includegraphics[width=\linewidth]{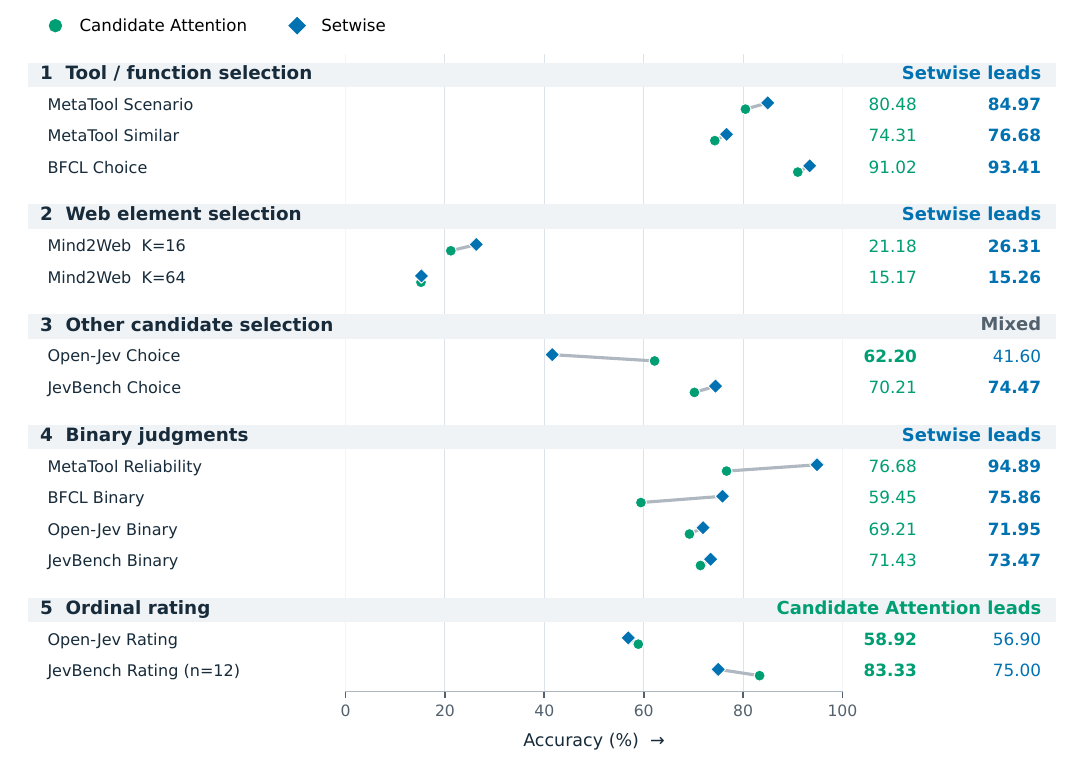}
        \par\smallskip
        {\small (b) RQ2: Candidate Attention vs.\ Setwise}
    \end{minipage}
    \hfill
    \begin{minipage}[t]{0.32\textwidth}
        \centering
        \includegraphics[width=\linewidth]{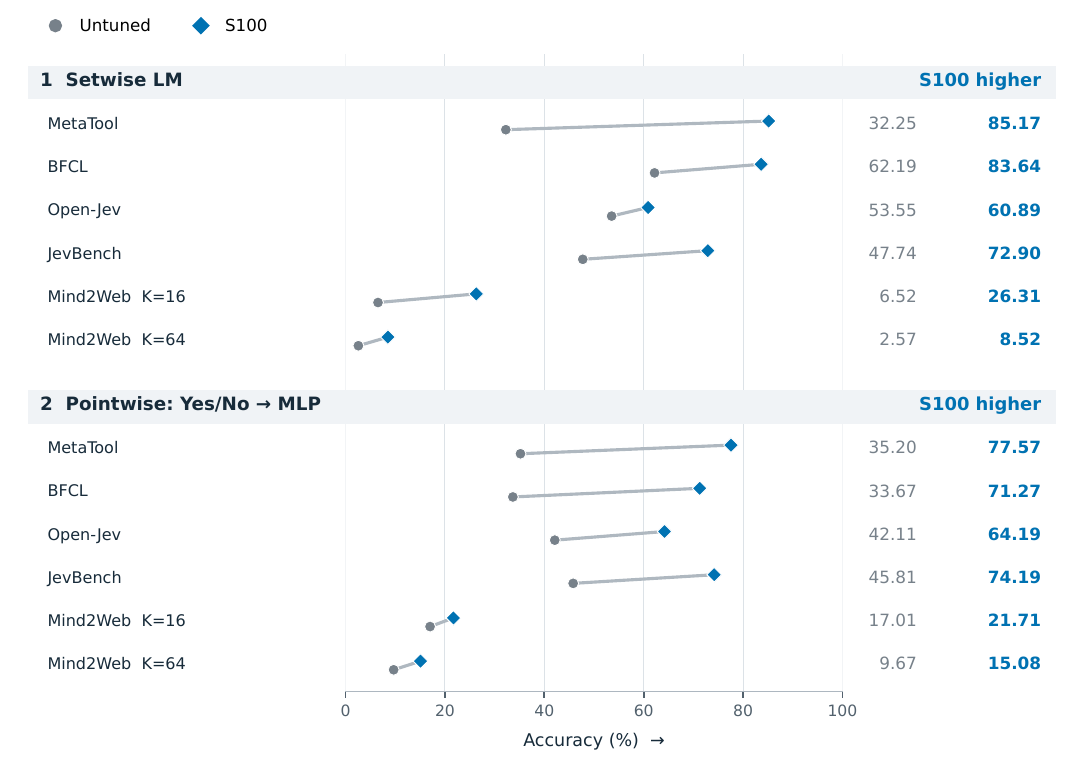}
        \par\smallskip
        {\small (c) RQ3: Transfer after S100 training}
    \end{minipage}

    \caption{Accuracy on agent decision benchmarks.
    (a) Fine-tuned pointwise and setwise methods.
    (b) Candidate Attention and fine-tuned setwise methods.
    (c) Untuned and S100 results.}
    \label{fig:rq_results}
\end{figure*}

We report results by research question. As S100 is our main training, condition Table~\ref{tab:benchmark_results} gives S100 accuracy on the five benchmarks, and Table~\ref{tab:synthetic_main} summarizes the paired synthetic tasks; complete results are in Appendix~\ref{app:results}. Our comparisons of scoring methods focus on fine-tuned models; we return to the untuned Setwise LM when
examining the effect of training in RQ3.

\subsection[RQ1: When Is Pointwise Scoring Sufficient?]{{RQ1: }When Is Pointwise Scoring Sufficient?}

\label{sec:rq1}


\paragraph{Answer to RQ1.}
We compare fine-tuned pointwise and setwise methods
under S100, reserving Candidate Attention for RQ2.
Figure~\ref{fig:rq_results}a reports the highest
observed accuracy within each family on each subtask.

Pointwise methods lead on other candidate selection
and ordinal rating, while setwise methods lead on
tool and function selection and binary judgments.
Web element selection shows a mixed pattern.

\paragraph{Where pointwise leads.}
On Open-Jev candidate selection, Pointwise MLP
achieves 62.60\% accuracy, compared with 41.60\%
for the best setwise method. Pointwise Linear also
leads on rating (62.29\% vs.\ 56.90\%).
JevBench shows a similar pattern, although its
subsets are small.

Further analysis suggests that part of this gap
comes from the scoring format. Pointwise methods
score each candidate directly, while Setwise LM
uses answer-label logits, which can be sensitive
to candidate positions. This is particularly
visible in Open-Jev, where some environments
reuse a small set of actions.

\paragraph{Where setwise leads.}
The best setwise methods outperform the best pointwise
methods on all three tool and function selection subtasks,
although each gap is below three percentage points.
For example, Setwise LM achieves 93.41\% on BFCL Choice,
compared with 91.14\% for Pointwise MLP.
The gap is larger on BFCL Binary:
Setwise LM reaches 75.86\%, while the best pointwise
method, Pointwise Linear, reaches 60.88\%.
Setwise LM also leads on web element selection with
$K=16$ (26.31\% versus 23.21\% retrieved accuracy). A closer analysis suggests that the pretrained
LM output head may contribute to these differences
through its conservative Yes/No predictions and
familiarity with multiple-choice formats.

Further analysis to RQ1 is provided in
Appendix~\ref{app:rq1_analysis}.

\subsection{RQ2: Does Candidate Attention Improve Decision Accuracy?}
\label{sec:rq2}

\paragraph{Answer to RQ2.}
Candidate Attention matches or exceeds Setwise methods on some tasks
but not consistently, and where it wins, the independent Pointwise
MLP wins by the same margin.
Under S100 we compare it with the better of Setwise LM and Setwise
Pointer on each subtask (Figure~\ref{fig:rq_results}b).

\paragraph{Where Candidate Attention performs well.}
It leads on Open-Jev choice (62.20\% vs.\ 41.60\%) and rating
(58.92\% vs.\ 56.90\%), and on JevBench rating, a subset of 12
questions.
At $K=64$ on web elements it is level with Setwise Pointer
(15.17\% vs.\ 15.26\%).

\paragraph{Where Candidate Attention falls behind.}
The best setwise methods lead on all tool and function subtasks and
all binary subtasks, on JevBench choice (70.21\% vs.\ 74.47\%), and
on web elements at $K=16$ (21.18\% vs.\ 26.31\%).
Both web comparisons use steps where a correct element is retrieved.

\paragraph{Does interaction explain its strong results?}
No.
Candidate Attention behaves like Pointwise MLP and does not
outperform it in general.

Both lead setwise scoring on the same Open-Jev sources.
On painting-geometry, where comparing candidates could help,
Candidate Attention is below Pointwise MLP (63.3\% vs.\ 71.1\%).
On MetaTool Similar, the pre-labeled comparative task, the two
differ by 0.1 points.

The main difference is on binary decisions, where Candidate
Attention gains 2--4 points by predicting \emph{Yes} less often
(34.7\% vs.\ 52.3\% on BFCL).
These decisions involve one candidate, so the gain reflects a
shifted threshold, not a benefit from comparing candidates. 

Further analysis to RQ2 is provided in
Appendix~\ref{app:rq2_analysis}.

\subsection[RQ3: Which Training Conditions Help, and Do They Generalize?]{RQ3: Which Training Conditions Help, and Do They Generalize?}
\label{sec:training_generalization}

\paragraph{Answer to RQ3.}
S100 training supports transfer to other agent decision
benchmarks (Figure~\ref{fig:rq_results}c).
Adding relational training further improves Setwise LM
on both synthetic tasks, including inputs that differ
from those used in training.
Candidate Attention shows some learning on familiar
inputs, but does not show the same transfer.

\paragraph{Source training transfers to agent benchmarks.}
S100 fine-tuning improves Setwise LM's accuracy on
all five benchmarks.
Overall accuracy rises from 32.25\% to 85.17\% on
MetaTool and from 62.19\% to 83.64\% on BFCL.
Trained pointwise methods also outperform untuned
Pointwise Yes/No. For example, Pointwise MLP reaches
64.19\% on Open-Jev, compared with 42.11\%.

\paragraph{Source training gives limited relational transfer.}
We next compare Setwise LM on the same synthetic
validation examples, using number ranges, tool functions,
and description templates absent from relational training.
S100 raises numeric median accuracy from 25.00\%
to 35.67\%, but semantic singleton accuracy falls
from 21.00\% to 10.00\%.
Thus, the gains on agent benchmarks do not imply
reliable performance on these relational tasks.

\paragraph{Adding relational training improves both tasks.}
S90R10 and S50R50 replace 10\% and 50\% of source
requests with relational tasks, respectively, while
keeping the total training budget fixed.
For Setwise LM, both mixtures achieve at least 99\%
accuracy on both tasks when held-out examples use
the training ranges, functions, and templates.

On the shifted validation set, S90R10 reaches
81.33\% numeric and 99.67\% semantic accuracy;
S50R50 reaches 82.67\% and 92.67\%.
Both mixtures therefore improve performance on new
inputs within these two task families. 

\section{Related Work}

\subsection{Candidate Scoring in Language Models}

Language-model-based systems have been used for various decision tasks,
including tool selection, function calling, and structured
or multiple-choice decisions
\citep{huang2024metatool,patil2025bfcl,zheng2024toolrerankadaptivehierarchyawarereranking,
yang-etal-2022-zero,typesafe2026jev,strands2026decider}.
Despite their different applications, these tasks often require selecting among candidates based on a given context.
Similar candidate scoring problems have been widely studied
in information retrieval, where pointwise methods score
candidates independently
\citep{nogueira-etal-2020-document, zhuang2022rankt5finetuningt5text},
while pairwise, listwise, and setwise methods consider multiple candidates
\citep{qin-etal-2024-large,sun-etal-2023-chatgpt,
Zhuang_2024}.

While these methods differ in candidate information
access, their comparisons often also involve changes
in prompt format, scoring readout, and candidate
ordering. Such differences can affect the results:
fine-grained relevance labels can improve pointwise
ranking \citep{zhuang-etal-2024-beyond}, while
candidate ordering and option identifiers can
influence predictions
\citep{tang-etal-2024-found, zheng2024large}.
Similar selection biases have been observed in
tool-use settings \citep{blankenstein2026biasbustersuncoveringmitigatingtool}.
Architectural comparisons can also involve multiple
changes: SBERT2S1 compares bi-encoder and cross-head
models for typed decisions that differ in both
state--candidate encoding and cross-candidate
interaction \citep{deka2026retrievaltypeddecisionscalibrated}.
We compare independent and interacting scoring
across diverse agent decision tasks, examining
how performance varies with scoring readout,
candidate ordering, and candidate-set characteristics.

\subsection{Context-Dependent Choice and Interaction Architectures}

Prior work has shown that a candidate's utility can depend on the other available choices. Context-dependent choice models such as FETA and FATE explicitly capture these dependencies \citep{Pfannschmidt_2022}, while context-aware learning-to-rank uses self-attention to incorporate information from other candidates \citep{pobrotyn2021contextawarelearningrankselfattention}. We further examine the role of candidate-set information in language-model-based agent decisions. 

To capture such dependencies, prior work has explored several interaction architectures. Deep Sets and Set Transformer provide general set-based architectures \citep{zaheer2018deepsets,lee2019settransformer}. In ranking, SetRank and Set-Encoder use cross-candidate attention \citep{pang2020setranklearningpermutationinvariantranking,Schlatt_2025}, while ListT5 combines independent passage encoding with joint decoding \citep{yoon-etal-2024-listt5}. Recent work on visual candidate selection finds limited benefits from cross-candidate interaction beyond request-conditioned encoding \citep{li2026seeing}. We extend prior work on candidate interaction in choice and ranking to agent decisions, comparing independent and interacting scoring methods. Our matched-head evaluation uses identical frozen candidate representations to isolate the effect of cross-candidate interaction.

\subsection{Learning and Generalization of Candidate Interactions}

Access to other candidates does not guarantee effective comparison, which may depend on the training signal. In passage reranking, Set-Encoder shows that standard relevance training may not fully exploit cross-passage interactions, while duplicate-aware supervision encourages their use \citep{Schlatt_2025}. Targeted training has
also been used in tool selection to help models
distinguish similar tools
\citep{mekala-etal-2024-toolverifier}.

Beyond improving selection on familiar candidates,
models must also handle unseen candidates and input
distributions. Prior work has studied generalization
to unseen tools \citep{he-etal-2025-gentool}, as well
as unseen relational combinations, numerical ranges,
and set sizes
\citep{sinha-etal-2019-clutrr,wang2020extrapolatablerelationalreasoningcomparators}.
However, it remains unclear whether relational training
changes the performance gap between independent and
interacting scoring, and whether this gap persists
under distribution shifts. We examine both questions
by comparing scoring methods under different training
mixtures and evaluating them on held-out and shifted
inputs.

\section{Conclusion}
\label{sec:conclusion}
We studied when decision models should compare
candidates rather than score them independently.
Across five benchmarks, Pointwise scoring remained
competitive, while Setwise scoring performed better
on some tool-selection and binary-decision tasks.
Candidate Attention offered no consistent gains
over Pointwise scoring. Controlled experiments
further showed that access to other candidates
does not guarantee effective comparison, even
with relational supervision.
\section*{Limitations}
Our study has several limitations. First, we evaluate only one 2B-parameter model and do not assess hardware-specific inference efficiency, despite prior work on heterogeneous CPU/GPU execution~\citep{alzantot2017rstensorflow}, while the deployment efficiency of different candidate-scoring methods remains outside the scope of this study. Second, we examine a limited set of
interaction architectures and two synthetic relational
tasks, which may not capture more complex candidate
dependencies. Finally, fixed training budgets,
input truncation, and small evaluation subsets
may affect the observed results.

\section*{Ethical Considerations}

This study uses existing benchmarks and synthetic data
without executing real-world actions. Incorrect agent
decisions may pose risks in deployment.

\section*{AI Assistance Disclosure}

GPT and Claude were used to assist with code development and language revision. The authors reviewed and verified the final code, experimental results, and manuscript content.
\bibliography{references,benchmark_references}
\appendix
\section{Implementation and Training Details}
\label{app:impl}

\paragraph{Backbone and adaptation.}
All fine-tuned methods start from Qwen3.5-2B-Base. We freeze the original language-model weights and
train LoRA adapters with rank 16, scaling 32, and dropout 0.05. The
adapters target the attention projections
(\texttt{q/k/v/o\_proj}), feed-forward projections
(\texttt{gate/up/down\_proj}), and the gated-delta projections used by
Qwen3.5 (\texttt{in\_proj\_qkv/z/a/b} and \texttt{out\_proj}). The original LM
output head remains frozen; learned scoring heads are trained jointly
with the adapters.

\paragraph{Optimization.}
We use AdamW with learning rates of $10^{-4}$ for LoRA parameters
and $10^{-3}$ for learned scoring heads, weight decay of 0.01,
and gradient clipping at a global norm of 1.
The learning rate follows a linear warmup for 108 steps (3\%
of training), followed by cosine decay.
Each of S100, S90R10, and S50R50 is trained for 3,606
optimization steps with an effective batch size of 32 requests,
totaling 115,392 training examples processed.
An additional diagnostic setting, R100-short, uses 500 steps
(16,000 examples processed).
We use the final checkpoint from each run without
validation-based checkpoint selection.

\paragraph{Training Hardware.} All experiments were conducted using 4 x NVIDIA
GeForce RTX 5090 GPUs, with an estimated total
compute budget of 200 GPU-hours.

\paragraph{Targets and precision.}
Choice and binary examples use hard gold targets. For an ordinal
rating, the gold level receives probability 0.9 and the remaining 0.1
is divided equally among its adjacent levels. The backbone and cached
features use BF16; learned heads and score normalization use FP32. The
joint LM-label readout retains the backbone's BF16 matrix multiplication
before FP32 normalization.

\paragraph{Candidate augmentation.}
Choice and binary candidates are shuffled at every training exposure
using a request-, step-, and seed-keyed permutation. Ordinal scales
retain their adjacency and are reversed with probability 0.5. Formal
evaluation uses the supplied candidate order unless an order control is
explicitly named.

\section{Prompts, Readouts, and Head Architecture}
\label{app:prompts}

All trained methods use the Qwen XML-style framing. The shared prefix
contains \texttt{<state>}$x$\texttt{</state>} followed by
\texttt{<question type="$d$">}$q$, where $d$ identifies a choice,
binary, or rating decision. Setwise methods append all candidates as
lettered options and the instruction ``Select exactly one option;
answer with its letter.'' Pointwise methods form one prompt per
candidate, append that candidate inside \texttt{<candidate>} tags, and
ask ``Is this candidate correct? Answer Yes or No.'' Candidate keys
used for auditing and tie-breaking are never included in pointwise
prompts.

Table~\ref{tab:head_configurations} specifies the candidate access,
readout, and head configuration for every architecture. Trainable head
counts exclude LoRA parameters. Setwise LM and Pointwise Yes/No reuse
the frozen LM output layer and therefore add no trainable head.

\begin{table*}[t]
\centering
\small
\setlength{\tabcolsep}{4pt}
\resizebox{\textwidth}{!}{%
\begin{tabular}{llllr}
\toprule
Architecture & Candidate access & Readout & Head configuration & Head params \\
\midrule
Setwise LM & Joint & Option-label logits & Frozen LM output layer & 0 \\
Setwise Pointer & Joint & Query--option dot product & 2,048-to-256 projections & 1,053,184 \\
Pointwise Yes/No & Independent & Normalized Yes/No logits & Frozen LM output layer & 0 \\
Pointwise MLP & Independent & Scalar score & Width 256; FFN width 1,024 & 1,056,257 \\
Pointwise Linear & Independent & Scalar score & Layer norm plus linear map & 6,145 \\
Pointwise Pointer & Independent & Query--candidate dot product & 2,048-to-256 projections & 1,053,184 \\
Candidate Attention & Interacting isolated states & Scalar score & Width 256; 4 heads; FFN width 512 & 1,057,281 \\
\bottomrule
\end{tabular}}
\caption{Architecture and scoring-head configurations. Candidate
Attention applies one self-attention block without candidate-position
embeddings. Pointwise MLP uses the same projection and scalar readout
but no attention. Pointwise MLP and Candidate Attention differ by 1,024
head parameters, or less than 0.1\%.}
\label{tab:head_configurations}
\end{table*}

Inputs are truncated independently of candidate order to 1,792 state
tokens, 192 question tokens, and 64 tokens per candidate. Training
examples contain 2--24 candidates and are asserted to fit within 4,096
tokens. For Mind2Web $K=64$, the verified single-token label inventory
is extended to 64 labels and the joint-input limit is 8,192 tokens. The
first 24 labels preserve their training mapping; labels 25--64 are
therefore a label-vocabulary extrapolation for Setwise LM. No candidate
is removed and a missing gold element is never inserted.

All six fine-tuned configurations are trained under S100 with seeds
0--2. The primary mixture comparison uses Pointwise MLP and Candidate
Attention under S90R10 and S50R50 with the same three seeds; Setwise LM
is also trained under both mixtures, with three S50R50 seeds and the
reported S90R10 seed-0 run. Table~\ref{tab:synthetic_results} reports
only architectures included in the unified synthetic re-evaluation;
an omitted method--condition combination was not evaluated or imputed.

Because Candidate Attention adds a newly initialized interaction block,
we also retrain its R100-short seed-1 model with head learning rates
$3\times10^{-4}$ and $3\times10^{-3}$, holding the data order, LoRA
learning rate, and all other settings fixed. These runs diagnose
optimization sensitivity; they are not a multiple-seed learning-rate
comparison.

\section{Controlled Comparison Details}
\label{app:controlled}

\paragraph{Matched complete methods.}
Within each condition and seed, Pointwise MLP and Candidate Attention
start from the same base checkpoint and receive the same candidate-local
prompts, gold targets, exposure slots, candidate permutations, and
number of optimizer updates. Their head capacities are matched as
described above. Each method nevertheless learns its own LoRA adapter.
The main $G$-$E$ contrast therefore compares two complete training
procedures; it does not by itself identify the causal contribution of
the attention block.

\paragraph{Frozen-encoder control.}
We isolate the head more narrowly by freezing the S90R10 seed-0 encoder
from each method in turn. For each frozen encoder, we cache identical
candidate representations and train a newly initialized MLP head and
attention head with the same S90R10 examples, schedule, and seed. The
four combinations appear in Table~\ref{tab:synthetic_results}. On both
encoders, changing the head moves request accuracy by at most 0.4 points
on ID and OOD development. Strict paired success is zero in every OOD
development comparison; on ID development it is zero except for 0.3\%
with the attention encoder and attention head. This control applies to
the tested frozen representations and optimization recipe; it does not
show that every possible interaction module is ineffective.

\section{Dataset Construction and Split Audit}
\label{app:data}

\paragraph{Source pool.}
After group isolation and filtering, the source mixture contains
115,366 training, 3,535 development, and 3,678 calibration requests.
The conversion removes 269 duplicate rows, 470 rows with conflicting
targets, and 21 rows whose source groups overlap an official holdout.
An exact comparison of normalized states against 149,487 converted
benchmark rows finds no overlap. This audit does not rule out
paraphrases, partial overlap, shared upstream material, or contamination
from backbone pretraining.

\paragraph{Agent-decision benchmarks.}
Every S100 method is evaluated on the same 9,156 converted requests:
3,054 MetaTool, 1,883 BFCL, 2,002 Open-Jev, 155 JevBench, and 2,062
Mind2Web candidate-pool instances. For Mind2Web, 935 of 1,312 sampled
steps contain an acceptable element at $K=16$ and 1,127 at $K=64$.
Retrieved accuracy conditions on these steps; all-step accuracy keeps
all 1,312 steps and counts retrieval misses as errors. The benchmark
test sets were inspected in earlier experimental rounds, so all transfer
results in this paper are exploratory rather than untouched
confirmation results.

\paragraph{Relational pool and evaluation splits.}
The relational training pool has 6,000 requests (3,000 pairs), divided
equally between numeric median and semantic singleton selection.
Candidate sets contain 3, 5, or 7 candidates. Both members of a pair
remain in the same split, and groups and generated entities are disjoint
across splits.

The ID development set contains 600 requests (300 pairs) newly sampled
from the training generator. The OOD development set contains another
600 requests (300 pairs), and the original OOD test contains 1,200
requests (600 pairs). Numeric fees are sampled from 1--999 in training,
1,100--2,099 in OOD development, and 3,300--4,299 in OOD test. The
semantic OOD splits also change capability categories and description
templates. A further 1,200-request sample from the OOD-test generator
was drawn during analysis; we call it the OOD resample rather than a
confirmation set. All four synthetic evaluation splits have been
inspected and are reported as exploratory.

\section{Candidate Order and Paired Metrics}
\label{app:order}

For prompted setwise baselines, sorting by stable candidate keys fixes
one serialization but can preserve a correlation between key order and
the answer. The cyclic-shift control evaluates up to four evenly spaced
rotations, maps probabilities back to candidate identity, and averages
them. It reduces dependence on one presentation order but does not
make a model invariant to arbitrary permutations. Pointwise raw scores
and Candidate Attention scores are permutation-equivariant by
construction; stable candidate keys resolve exact final-score ties.

For a pair $p$ with decisions $(u_p,v_p)$, raw paired success is
\[
  \frac{1}{P}\sum_{p=1}^{P}
  \mathbf{1}[\hat y_{u_p}=y_{u_p}\ \land\
              \hat y_{v_p}=y_{v_p}].
\]
Historical archived tables use the fixed candidate-key tie rule for raw
paired success. The unified synthetic tables below instead report
strict Pair, which requires the gold score to exceed every alternative
by more than $10^{-4}$ in both members of the pair. Tiny nonzero raw
paired values from an independent scorer can arise from exact or near
ties and BF16 execution differences; they are not evidence that the
scorer used the surrounding candidates. We therefore distinguish the
historical raw metric from the strict metric throughout.

\section{Full Results and Statistical Details}
\label{app:results}

Table~\ref{tab:synthetic_main} gives the compact comparison used in the
main text. Table~\ref{tab:synthetic_results} expands it across all four
synthetic splits and reports seed variability where available.
Table~\ref{tab:mixture_benchmarks} reports exploratory transfer under
S100 and S50R50. Means use seeds 0--2 and error terms are sample
standard deviations; a labeled seed is a single trained checkpoint, not
an uncertainty estimate. No multiple-comparison correction is applied
to these descriptive tables.
\begin{table*}[t]
\centering
\small
\setlength{\tabcolsep}{2pt}
\begin{tabular}{ll*{5}{r}}
\toprule
& & \multicolumn{2}{c}{ID dev} & \multicolumn{2}{c}{OOD dev} & Source \\
\cmidrule(lr){3-4}\cmidrule(lr){5-6}\cmidrule(lr){7-7}
Training & Method & Acc & Pair & Acc & Pair & Acc \\
\midrule
Untuned & Setwise LM & 23.5 & 7.7 & 23.0 & 5.7 & 38.7 \\
\midrule
S100 & Setwise LM$^\dagger$ & 25.8 & 3.3 & 22.5 & 6.7 & 85.9 \\
 & Pointwise MLP & 12.9 & 0.0 & 21.8 & 0.0 & 85.6 \\
 & Cand.\ Attn. & 11.9 & 0.0 & 18.5 & 0.0 & 85.5 \\
\midrule
S90R10 & Setwise LM$^\dagger$ & 99.7 & 99.3 & 90.0 & 83.3 & 85.2 \\
 & Pointwise MLP & 33.7 & 0.0 & 16.9 & 0.0 & 85.2 \\
 & Cand.\ Attn. & 33.1 & 0.0 & 19.8 & 0.0 & 85.3 \\
\midrule
S50R50 & Setwise LM & 99.7 & 99.3 & 89.8 & 81.0 & 84.3 \\
 & Pointwise MLP & 33.9 & 0.0 & 17.2 & 0.0 & 84.3 \\
 & Cand.\ Attn. & 40.4 & 8.6 & 20.4 & 0.0 & 84.4 \\
\midrule
R100-short & Setwise LM & 98.9 & 97.7 & 93.3 & 87.7 & 58.4 \\
 & Pointwise MLP$^\dagger$ & 33.0 & 0.0 & 22.0 & 0.0 & 26.5 \\
 & Cand.\ Attn. & 46.8 & 19.2 & 20.7 & 0.0 & 26.6 \\
\midrule
Frozen$^\ddagger$ & MLP head & 34.0 & 0.0 & 19.2 & 0.0 & 85.6 \\
 & Attn.\ head & 33.8 & 0.0 & 18.8 & 0.0 & 85.6 \\
\bottomrule
\end{tabular}
\caption{Synthetic relational summary, in percent. ID/OOD values use
 the unified re-evaluation protocol of Table~\ref{tab:synthetic_results}.
 Acc is request accuracy; Pair requires the gold log-probability to
 exceed every alternative by more than $10^{-4}$ in both pair members.
 ID follows the training generator; OOD dev changes numeric ranges,
 semantic capabilities, and templates. Trained-model values are means
 over seeds 0--2 unless marked $^\dagger$ (seed 0).
 $^\ddagger$ Newly trained seed-0 heads on the frozen S90R10 seed-0
 Pointwise MLP encoder. Source reports archived accuracy on the 3,535
 source-development requests for the corresponding checkpoints; it was
 not rerun in the synthetic re-evaluation. Evaluation hardware and
 numerical caveats are detailed in Table~\ref{tab:synthetic_results}.
 All synthetic splits are exploratory.}
\label{tab:synthetic_main}
\end{table*}

\begin{table*}[t]
\centering
\small
\setlength{\tabcolsep}{4pt}
\resizebox{\textwidth}{!}{%
\begin{tabular}{l*{8}{r}}
\toprule
& \multicolumn{2}{c}{ID dev}
& \multicolumn{2}{c}{OOD dev}
& \multicolumn{2}{c}{OOD test}
& \multicolumn{2}{c}{OOD resample$^\ast$} \\
\cmidrule(lr){2-3}\cmidrule(lr){4-5}\cmidrule(lr){6-7}\cmidrule(lr){8-9}
Method & Acc & Pair & Acc & Pair & Acc & Pair & Acc & Pair \\
\midrule
\multicolumn{9}{l}{\textit{No task-specific training}} \\
Setwise LM & 23.5 & 7.7 & 23.0 & 5.7 & 22.2 & 6.7 & 23.7 & 6.2 \\
\midrule
\multicolumn{9}{l}{\textit{S100}} \\
Setwise LM (seed 0) & 25.8 & 3.3 & 22.5 & 6.7 & 27.9 & 7.5 & 26.2 & 8.2 \\
Pointwise MLP & 12.9\,{\tiny$\pm$0.8} & 0.0\,{\tiny$\pm$0.0} & 21.8\,{\tiny$\pm$4.1} & 0.0\,{\tiny$\pm$0.0} & 25.6\,{\tiny$\pm$4.5} & 0.0\,{\tiny$\pm$0.0} & 24.6\,{\tiny$\pm$4.6} & 0.0\,{\tiny$\pm$0.0} \\
Candidate Attention & 11.9\,{\tiny$\pm$1.1} & 0.0\,{\tiny$\pm$0.0} & 18.5\,{\tiny$\pm$8.4} & 0.0\,{\tiny$\pm$0.0} & 26.8\,{\tiny$\pm$3.2} & 0.0\,{\tiny$\pm$0.0} & 25.4\,{\tiny$\pm$4.4} & 0.0\,{\tiny$\pm$0.0} \\
\quad head LR $3\times10^{-4}$ (seed 0) & 11.8 & 0.0 & 20.3 & 0.0 & 30.2 & 0.0 & 27.8 & 0.0 \\
\midrule
\multicolumn{9}{l}{\textit{S90R10}} \\
Setwise LM (seed 0) & 99.7 & 99.3 & 90.0 & 83.3 & 89.5 & 80.2 & 88.5 & 78.8 \\
Setwise Pointer (seed 0) & 65.2 & 45.0 & 60.8 & 36.7 & 48.5 & 24.0 & 45.3 & 22.0 \\
Pointwise MLP & 33.7\,{\tiny$\pm$0.4} & 0.0\,{\tiny$\pm$0.0} & 16.9\,{\tiny$\pm$4.2} & 0.0\,{\tiny$\pm$0.0} & 17.2\,{\tiny$\pm$11.4} & 0.0\,{\tiny$\pm$0.0} & 16.0\,{\tiny$\pm$11.0} & 0.0\,{\tiny$\pm$0.0} \\
Pointwise Linear (seed 0) & 33.3 & 0.0 & 11.0 & 0.0 & 12.2 & 0.0 & 13.5 & 0.0 \\
Candidate Attention & 33.1\,{\tiny$\pm$0.9} & 0.0\,{\tiny$\pm$0.0} & 19.8\,{\tiny$\pm$2.3} & 0.0\,{\tiny$\pm$0.0} & 16.3\,{\tiny$\pm$9.0} & 0.0\,{\tiny$\pm$0.0} & 16.7\,{\tiny$\pm$7.4} & 0.0\,{\tiny$\pm$0.0} \\
Pointwise Pointer (seed 0) & 34.0 & 0.0 & 28.8 & 0.0 & 27.0 & 0.0 & 27.5 & 0.0 \\
\midrule
\multicolumn{9}{l}{\textit{S50R50}} \\
Setwise LM & 99.7\,{\tiny$\pm$0.2} & 99.3\,{\tiny$\pm$0.3} & 89.8\,{\tiny$\pm$2.7} & 81.0\,{\tiny$\pm$5.0} & 91.4\,{\tiny$\pm$2.0} & 83.9\,{\tiny$\pm$3.9} & 89.0\,{\tiny$\pm$2.6} & 79.8\,{\tiny$\pm$4.9} \\
Setwise Pointer (seed 0) & 99.7 & 99.3 & 80.7 & 66.3 & 63.2 & 47.8 & 60.6 & 43.3 \\
Pointwise MLP & 33.9\,{\tiny$\pm$1.4} & 0.0\,{\tiny$\pm$0.0} & 17.2\,{\tiny$\pm$9.0} & 0.0\,{\tiny$\pm$0.0} & 19.1\,{\tiny$\pm$6.9} & 0.0\,{\tiny$\pm$0.0} & 19.6\,{\tiny$\pm$6.1} & 0.0\,{\tiny$\pm$0.0} \\
Pointwise Linear (seed 0) & 35.2 & 0.0 & 14.5 & 0.0 & 11.8 & 0.0 & 11.8 & 0.0 \\
Candidate Attention & 40.4\,{\tiny$\pm$10.2} & 8.6\,{\tiny$\pm$13.4} & 20.4\,{\tiny$\pm$5.6} & 0.0\,{\tiny$\pm$0.0} & 24.3\,{\tiny$\pm$5.9} & 0.0\,{\tiny$\pm$0.0} & 22.8\,{\tiny$\pm$4.6} & 0.0\,{\tiny$\pm$0.0} \\
\quad head LR $3\times10^{-4}$ (seed 0) & 87.8 & 79.3 & 19.3 & 3.3 & 43.4 & 13.5 & 44.4 & 13.8 \\
Pointwise Pointer (seed 0) & 33.8 & 0.0 & 26.0 & 0.0 & 25.2 & 0.0 & 23.1 & 0.0 \\
\midrule
\multicolumn{9}{l}{\textit{R100-short (500 updates, relational data only)}} \\
Setwise LM & 98.9\,{\tiny$\pm$0.3} & 97.7\,{\tiny$\pm$0.9} & 93.3\,{\tiny$\pm$2.1} & 87.7\,{\tiny$\pm$4.1} & 93.6\,{\tiny$\pm$1.2} & 88.1\,{\tiny$\pm$2.3} & 93.1\,{\tiny$\pm$1.7} & 87.1\,{\tiny$\pm$2.7} \\
Pointwise MLP (seed 0) & 33.0 & 0.0 & 22.0 & 0.0 & 6.6 & 0.0 & 6.9 & 0.0 \\
Candidate Attention & 46.8\,{\tiny$\pm$20.2} & 19.2\,{\tiny$\pm$29.3} & 20.7\,{\tiny$\pm$1.8} & 0.0\,{\tiny$\pm$0.0} & 11.5\,{\tiny$\pm$8.6} & 0.0\,{\tiny$\pm$0.0} & 10.9\,{\tiny$\pm$8.8} & 0.1\,{\tiny$\pm$0.1} \\
\quad head LR $3\times10^{-4}$ (seed 1) & 87.8 & 77.3 & 28.3 & 1.3 & 20.0 & 1.5 & 19.6 & 0.3 \\
\quad head LR $3\times10^{-3}$ (seed 1) & 33.0 & 0.3 & 27.3 & 0.0 & 9.9 & 0.0 & 9.0 & 0.0 \\
\midrule
\multicolumn{9}{l}{\textit{Frozen S90R10 seed-0 encoder, new head (seed 0)}} \\
MLP encoder + MLP head & 34.0 & 0.0 & 19.2 & 0.0 & 11.8 & 0.0 & 10.9 & 0.0 \\
MLP encoder + attention head & 33.8 & 0.0 & 18.8 & 0.0 & 11.8 & 0.0 & 10.9 & 0.0 \\
Attention encoder + MLP head & 32.2 & 0.0 & 20.3 & 0.0 & 11.6 & 0.0 & 13.3 & 0.0 \\
Attention encoder + attention head & 32.3 & 0.3 & 20.5 & 0.0 & 11.5 & 0.0 & 13.0 & 0.0 \\
\bottomrule
\end{tabular}
}
\caption{Synthetic relational evaluation under the unified re-evaluation
protocol. Acc is request accuracy; Pair is strict paired success:
both members must have a gold-versus-best-alternative log-probability
margin greater than $10^{-4}$. Values are percentages, pooling numeric
median and semantic singleton examples. Means and sample standard
deviations use seeds 0--2; single trained checkpoints are labeled.
Identical candidate-local inputs reuse their encoded representations;
independent scorers also reuse their scalar scores. Stable candidate
keys resolve accuracy ties. All four splits are exploratory.
$^\ast$A second sample from the OOD-test generator, inspected during
analysis.}
\label{tab:synthetic_results}
\end{table*}

\begin{table*}[t]
\centering
\small
\setlength{\tabcolsep}{4pt}
\resizebox{\textwidth}{!}{
\begin{tabular}{l*{8}{r}}
\toprule
& MetaTool & BFCL
& \multicolumn{2}{c}{Mind2Web, $K=16$}
& \multicolumn{2}{c}{Mind2Web, $K=64$}
& Open-Jev & JevBench \\
\cmidrule(lr){4-5}\cmidrule(lr){6-7}
Method & Overall & Overall & Retrieved & All & Retrieved & All & Overall & Overall \\
\midrule
\multicolumn{9}{l}{\textit{S100, seed 0}} \\
Setwise LM (fine-tuned) & 85.17 & 83.64 & 26.31 & 18.75 & 8.52 & 7.32 & 60.89 & 72.90 \\
Pointwise MLP & 77.57 & 71.27 & 21.71 & 15.47 & 15.08 & 12.96 & 64.19 & 74.19 \\
Candidate Attention & 77.87 & 73.45 & 21.18 & 15.09 & 15.17 & 13.03 & 65.93 & 71.61 \\
\midrule
\multicolumn{9}{l}{\textit{S50R50, mean $\pm$ std over seeds 0--2}} \\
Setwise LM (fine-tuned) & 85.30\,{\tiny$\pm$0.87} & 79.25\,{\tiny$\pm$0.46} & 24.24\,{\tiny$\pm$0.31} & 17.28\,{\tiny$\pm$0.22} & 6.98\,{\tiny$\pm$0.36} & 6.00\,{\tiny$\pm$0.31} & 59.56\,{\tiny$\pm$1.19} & 72.90\,{\tiny$\pm$2.32} \\
Pointwise MLP & 83.64\,{\tiny$\pm$1.10} & 77.32\,{\tiny$\pm$2.35} & 20.68\,{\tiny$\pm$2.20} & 14.74\,{\tiny$\pm$1.57} & 13.37\,{\tiny$\pm$1.52} & 11.48\,{\tiny$\pm$1.31} & 61.16\,{\tiny$\pm$0.92} & 69.46\,{\tiny$\pm$0.75} \\
Candidate Attention & 83.57\,{\tiny$\pm$1.18} & 77.66\,{\tiny$\pm$2.79} & 21.82\,{\tiny$\pm$1.42} & 15.55\,{\tiny$\pm$1.01} & 13.93\,{\tiny$\pm$0.55} & 11.97\,{\tiny$\pm$0.48} & 60.82\,{\tiny$\pm$0.59} & 70.33\,{\tiny$\pm$1.29} \\
\bottomrule
\end{tabular}
}
\caption{Exploratory benchmark accuracy (\%) under S100 and S50R50.
S100 reports one checkpoint; S50R50 reports the mean and sample standard
deviation over three seeds. Mind2Web Retrieved conditions on a pool
containing an acceptable element, whereas All counts retrieval failures
as errors. These test sets had been inspected before the mixture runs.}
\label{tab:mixture_benchmarks}
\end{table*}

\section{RQ1 Analysis: Readout Effects behind Family Gaps}
\label{app:rq1_analysis}

This section diagnoses the S100 seed-0 differences summarized in
Figure~\ref{fig:rq_results}a. The comparisons change candidate
visibility, prompt format, and readout together, so the observations
below identify plausible readout mechanisms rather than causal effects
of visibility.

\subsection{Label preference and content preference}

Setwise LM scores answer labels (A, B, C, \ldots), whereas pointwise
heads score candidate representations directly. Open-Jev makes this
difference visible because several environments reuse a small action
inventory across many steps.

\paragraph{Tile platformer.}
Every evaluated step offers the same seven actions:
\texttt{noop}, \texttt{right}, \texttt{right\_jump},
\texttt{right\_run}, \texttt{right\_run\_jump}, \texttt{jump}, and
\texttt{left}. The gold action is \texttt{right\_run\_jump} on nearly
all 105 steps. The untuned Setwise LM selects the label mapped to
\texttt{left} on every step; the fine-tuned Setwise LM alternates between
two other labels and never selects the gold action; Setwise Pointer also
selects it on none. Each pointwise method selects it on 89.5\% of
steps. This pattern is consistent with a reusable preference for the
candidate content in the pointwise heads and a persistent label-position
preference in the setwise readout.

This one nearly constant-target source explains much of the 21-point
family gap on Open-Jev choice. The pointwise lead is smaller when the
gold action varies: on painting-geometry, with about twelve candidates
per step, Pointwise MLP reaches 71.1\% and Setwise LM 61.7\%. On snake,
Setwise LM reaches 49.3\%, below the untuned setwise model's 62.0\%.
These cases show that source fine-tuning can alter the label readout in
ways that do not transfer uniformly across action formats.

\paragraph{Ordinal ratings.}
For a rating task, Setwise LM sees labels attached to the levels and
answers with a letter. The label itself carries no ordinal geometry,
and its meaning changes with scale length: label D is the maximum on a
four-level row but the second-highest level on a five-level row. A
pointwise head instead receives the level text in each candidate prompt,
and ordinal target smoothing gives adjacent levels related supervision.
The observed advantage concentrates on vizdoom (Pointwise Linear 81.8\%
versus Setwise LM 39.4\%) and reverses on reasoning-control (Pointwise
MLP 37.5\% versus Setwise LM 55.0\%), so it is not a universal property
of rating tasks.

\subsection{Binary thresholds}

A binary gate is represented by affirmative and negative leaves but
does not require choosing among multiple domain candidates. Setwise LM
retains the pretrained output layer, whereas the pointwise scalar heads
learn a new threshold. Table~\ref{tab:gate_yes_rate} reports how often
each method predicts \emph{Yes}.

\begin{table*}[t]
\centering
\small
\setlength{\tabcolsep}{4pt}
\resizebox{\textwidth}{!}{%
\begin{tabular}{lrrrrrrrr}
\toprule
& & & \multicolumn{5}{c}{Predicted Yes (\%)} & \\
\cmidrule(lr){4-8}
Dataset & $n$ & Gold Yes (\%)
& Untuned & Setwise LM & MLP (E) & Linear (F)
& Attention (G) & G$-$E Acc. \\
\midrule
BFCL & 1,048 & 16.5 & 0.0 & 26.1 & 52.3 & 39.0 & 34.7 & $+4.0$ \\
Open-Jev & 1,205 & 30.0 & 0.0 & 30.5 & 38.4 & 38.3 & 37.5 & $+3.6$ \\
MetaTool reliability & 802 & 0.0 & 0.0 & 5.1 & 27.6 & 12.8 & 23.3 & $+4.2$ \\
JevBench & 49 & 42.9 & 0.0 & 61.2 & 57.1 & 57.1 & 63.3 & $+2.0$ \\
\bottomrule
\end{tabular}
}
\caption{Binary-decision diagnostics on the exploratory transfer
benchmarks. Gold and predicted Yes rates are percentages; G$-$E Acc.\
is the accuracy difference in percentage points. Untuned denotes the
original Setwise LM; all trained methods use S100, seed 0. Differences
in positive prediction rates describe decision behavior but do not
establish the cause of the accuracy differences.}
\label{tab:gate_yes_rate}
\end{table*}

Only 16.5\% of BFCL targets are positive. Setwise LM predicts
\emph{Yes} on 26.1\% and reaches 75.9\% accuracy; Pointwise MLP predicts
\emph{Yes} on 52.3\% and reaches 55.4\%. On JevBench, whose positive
rate is 42.9\%, the gap narrows to 73.5\% versus 69.4\%. MetaTool
reliability contains no positive targets, so its accuracy is exactly one
minus the false-positive rate. These results support a threshold-transfer
explanation, but the experiment does not isolate whether prevalence,
prompt format, or the frozen LM output head produces the threshold.

\subsection{Tool choice and candidate count}

Both families are strong on the three tool and function choice subtasks:
BFCL Choice is 93.4\% for Setwise LM and 91.1\% for Pointwise MLP;
MetaTool scenario is 85.0\% for Setwise Pointer and 82.3\% for Pointwise
MLP; MetaTool similar is 76.7\% for Setwise LM and 74.8\% for Pointwise
Linear. A two- to three-point edge can reflect normalization over the
full tool list, the suitability of the frozen multiple-choice output
layer, or candidate interaction. The matched E/G comparison and the
frozen-encoder control do not show a corresponding interaction gain, so
the present evidence does not single out the last mechanism.

On Mind2Web, Setwise LM leads the best pointwise model at $K=16$ (26.3\%
versus 23.2\% retrieved accuracy) and trails at $K=64$ (8.5\% versus
16.4\%). The $K=64$ setwise run also uses 40 answer labels absent from
fine-tuning. Candidate count, retrieval difficulty, and label coverage
therefore change together; this comparison cannot attribute the drop to
candidate interaction or to label extrapolation alone.

\section{RQ2 Analysis: What Candidate Attention Changes}
\label{app:rq2_analysis}

Candidate Attention (G) and Pointwise MLP (E) use matched data,
candidate-local prompts, LoRA budgets, and head capacity. G adds one
four-head self-attention block over the candidates and narrows its
feed-forward block to preserve parameter count. Because the models learn separate LoRA adapters, the E/G contrasts derived from Table~\ref{tab:benchmark_results} compare complete training procedures. Appendix~\ref{app:controlled} is the head-only control.

\subsection{Choice behavior}

G and E differ by at most 0.5 points on five of seven displayed choice
subtasks. The exception with the largest magnitude is JevBench choice
($n=94$, $-5.3$ points); under S50R50, the three seed-specific $G$-$E$
differences range from 0.0 to $+4.3$, so the sign does not persist across
conditions and seeds.

Where G exceeds setwise scoring, E already does. On Open-Jev choice, G
leads the best setwise method by 20.6 points and E by 21.0. Both select
the same actions on tile platformer and snake (89.5\% and 69.0\%). On
painting-geometry, where candidates differ in fine geometric details,
G is below E (63.3\% versus 71.1\%). On MetaTool similar, the task with
the clearest competing tool descriptions, G and E differ by 0.1 points
under S100; their S50R50 seed differences are $+0.4$, $-0.1$, and
$-0.1$. These results show no consistent choice gain from the tested
interaction head.

At $K=16$ and $K=64$ on Mind2Web, S100 $G$-$E$ is $-0.5$ and $+0.1$
points. Under S50R50, the three-seed ranges are $-0.5$ to $+1.8$ and
$-0.6$ to $+1.1$. Increasing the number of candidates does not reveal a
larger advantage for this attention head.

\subsection{Binary behavior}

G exceeds E on all four binary subtasks by two to four points. The gain
coincides with a different positive rate (Table~\ref{tab:gate_yes_rate}).
Attention lowers E's \emph{Yes} rate on the three larger, negative-heavy
sets; on the small JevBench subset it raises the positive rate and still
improves accuracy. Because a gate compares affirmative and negative
leaves rather than multiple domain candidates, this result is evidence
of a changed binary decision boundary, not a demonstrated benefit on
candidate selection.

\subsection{Relational learning and scope}

The synthetic tasks separate representational capacity from observed
benchmark accuracy. Under S50R50, Candidate Attention reaches 8.6\%
mean strict paired success on ID development but 0\% on OOD development;
Pointwise MLP remains at 0\% on both, as expected for an independent
scorer. R100-short raises Candidate Attention to 19.2\% ID strict paired
success on average, with a 29.3-point standard deviation, while its OOD
mean remains at the numerical floor. Lowering the seed-1 head learning
rate raises ID strict paired success to 77.3\% but only 1.3\% OOD. The
tested head can therefore fit some in-distribution relations, but this
evidence does not show reliable transfer. Setwise LM learns both tasks
far more consistently under the same data conditions.

All S100 benchmark contrasts in this section use one seed and previously
inspected test sets. The S50R50 ranges contain only three trained seeds,
and the learning-rate variants contain one. The conclusions apply to
this backbone, head, optimization recipe, and data construction.

\end{document}